 \documentclass[preprint, review,3p,times,10pt]{elsarticle}

\usepackage{hyperref}
\usepackage{amsmath,amssymb,amsfonts}
\usepackage{subcaption}
\usepackage{array}
\usepackage{setspace}
\usepackage{multirow}
\usepackage{makecell}

\begin{document}
\doublespacing                    

\begin{frontmatter}



\title{MHAFF: Multi-Head Attention Feature Fusion of CNN and Transformer for Cattle Identification}


\author[inst1,inst2,inst3]{Rabin Dulal}\ead{rdulal@csu.edu.au}
\author[inst1,inst2,inst3]{Lihong Zheng}\ead{lzheng@csu.edu.au}
\author[inst1,inst2,inst3]{Muhammad Ashad Kabir}\ead{akabir@csu.edu.au}
\affiliation[inst1]{organization={School of Computing, Mathematics and Engineering, Charles Sturt University},
            state={NSW},
            country={Australia}}

\affiliation[inst2]{organization={Food Agility CRC Ltd},
            city={Sydney},
            postcode={2000}, 
            state={NSW},
            country={Australia}}
\affiliation[inst3]{organization={Gulbali Institute for Agriculture, Water and Environment},
            city={Wagga Wagga},
            postcode={2650}, 
            state={NSW},
            country={Australia}}

\begin{abstract}

Convolutional Neural Networks (CNNs) have drawn researchers' attention to identifying cattle using muzzle images. However, CNNs often fail to capture long-range dependencies within the complex patterns of the muzzle. The transformers handle these challenges. This inspired us to fuse the strengths of CNNs and transformers in muzzle-based cattle identification. Addition and concatenation have been the most commonly used techniques for feature fusion. However, addition fails to preserve discriminative information, while concatenation results in an increase in dimensionality. Both methods are simple operations and cannot discover the relationships or interactions between fusing features. This research aims to overcome the issues faced by addition and concatenation. This research introduces a novel approach called Multi-Head Attention Feature Fusion (MHAFF) for the first time in cattle identification. MHAFF captures relations between the different types of fusing features while preserving their originality. The experiments show that MHAFF outperformed addition and concatenation techniques and the existing cattle identification methods in accuracy on two publicly available cattle datasets. MHAFF demonstrates excellent performance and quickly converges to achieve optimum accuracy of 99.88\% and 99.52\% in two cattle datasets simultaneously.
\end{abstract}




\end{frontmatter}

\section{Introduction}
Increased biosecurity measures and food safety protocols are boosting the demand for efficient cattle traceability, achieved through accurate and effective identification systems. Cattle identification can broadly be classified into classical, electronic, and recent vision-based methods~\citep{LR1awad2016classical, hossain2022systematic}. Classical methods like ear tattooing, branding, ear tagging, and ear notching~\citep{CIOruiz2011role, LR1awad2016classical, INneary2002methods} have been used. The ear tag typically includes a unique identification number and other information, such as the animal's breed, sex, and date of birth. The tag is usually inserted into the animal's ear using a special applicator. Electronic methods are currently being applied and are more efficient than classical methods. Currently, cattle in Australia are identified through their National Livestock Identification System (NLIS) tags~\citep{NLIS}, which use radio frequency identification (RFID) technology. These tags offer great benefits in cattle identification. However, they are susceptible to attacks, and being lost, damaged, and altered~\citep{CIOruiz2011role, INneary2002methods, LR1awad2016classical, andrew2019visual, ruiz2011role, wang2010rfid}. Other biometric identification methods include iris patterns, retina imaging, DNA sequencing, coat patterns, and muzzle patterns. Retina and iris imaging technology require a lot of effort to capture images. DNA sequencing requires specialized labs, equipment, and a workforce, and getting a report is a lengthy process. The coat pattern works only for cattle with a distinct color pattern and cannot be applied to cattle with plain body color. 

Muzzle patterns are unique biometric identifiers like fingerprints in humans~\citep{kumar2020cattle}. The muzzle has a distinct dermatoglyphic pattern of beads and ridges~\citep{baranov1993breed}. The muzzle patterns enlarge with age, but their unique combination remains unchanged~\citep{petersen1922identification, LR1awad2016classical}. Muzzle patterns are easier to capture with photos without requiring laborious work and do not need specialized equipment or labs~\citep{AID32kumar2018deep, awad2019bag, kumar2017muzzle}. This makes the application of the muzzle pattern an effective and efficient biometric approach for cattle identification~\citep{CIOruiz2011role, INneary2002methods, LR1awad2016classical, andrew2019visual, ruiz2011role, wang2010rfid}.

In the past decade, Convolutional Neural Networks (CNNs) have been a very popular method in modern computer vision tasks~\citep{hou2015convolutional, sharif2014cnn} because of their capacity to extract meaningful features directly from the image data~\citep{lecun2015deep}. Similarly, CNNs are mostly applied in muzzle-based cattle identification and have shown great performance~\citep{AID32kumar2018deep, vincent2008extracting, bengio2009learning, bengio2013representation, shojaeipour2021automated, li2022individual, lee2023identification}. However, CNNs limit the capacity to capture long-range dependencies of the features~\citep{yang2024feature}. CNNs fail in capturing the full range of intricate patterns in the input data due to their limited receptive fields~\citep{yang2024feature, naseer2021intriguing}. Meanwhile, transformers inherently incorporate global information through their self-attention mechanism but lack inductive bias~\citep{dosovitskiy2020image}. This limits capturing detailed, fine-grained features (local features)~\citep{han2022survey, khan2022transformers} with transformer. Thus, only one type of feature limits overall performance and broader generalization~\citep{mogan2024ensemble}. To address this, integrating CNNs and transformers provides the combined advantages of both approaches~\citep{yang2024feature, li2023x, ojala1994performance, li2023combining, du2022individual, wan2023sheep, fu2022lightweight, hu2020cow, weng2022cattle}.


In cattle identification, features of CNNs and transformers are fused by using simple addition and concatenation methods~\citep{ojala1994performance, li2023combining, du2022individual, wan2023sheep, fu2022lightweight, hu2020cow, weng2022cattle}. Those researches have shown improved accuracy in the identification. For features \(X\) and \(Y\), the fusion by addition is represented as $X + Y$. The addition assumes that the combined features can be summed up. However, this may lead to the loss of discriminative information if the features have different scales, magnitudes, or semantics. Features with larger values may dominate and overshadow the contributions of features with smaller values~\citep{dai2021attentional}. Another popular method, concatenation, is denoted as \([X, Y]\), which combines features by appending them to each other. However, all features have been considered equally, and it generates a high dimensional vector, increasing the computational load~\citep{dai2021attentional, wei2020f3net}. These simple fusion methods may not adapt well to the varying nature of the information present in different parts of the network, potentially limiting the network's ability to capture and utilize complex and diverse relationships within the data because of the simple linear operations~\citep{Zhou_2022, Ding_2023, Dai_2020}. Addition and concatenation both perform their operations without knowing any relations between features. Thus, there is considerable opportunity to enhance cattle identification by utilizing a dynamic and context-aware feature fusion method.

In this research, we identified the cattle using muzzle images by leveraging the strength of CNN and transformer using a new feature fusion method. The major contributions of this research are as follows:
\begin{itemize}
    \item We propose a new feature fusion method that improves accuracy over addition and concatenation. The innovation of the current study is the new feature fusion method utilizing multi-head attention to make a context-aware feature fusion between CNN and transformer features. This work fuses local and global features, and the proposed method is validated using benchmark and cattle muzzle datasets. 
    \item We investigate the optimal performance of fused features by combining the Query, Key, and Value components of multi-head attention. We used transformer features such as Query and Value and CNN features as Key for the multi-head attention. This feature fusion method is applied for the first time in cattle identification. 
    \item We compare our proposed method with the existing muzzle-based cattle identification methods. Our method outperforms the existing cattle identification methods.
\end{itemize}

\section{Background}
This section describes the basic concepts of deep networks, particularly CNN and transformer networks, which were used in this study. 

\subsection{CNN}
CNNs are designed for spatially local processing. Convolutional layers use small filters (e.g., $3\times3$, $5\times5$) to capture local patterns and hierarchies in the input, called local features~\citep{li2021survey}. This subsection provides a brief introduction to the CNNs used in this study.

VGG16~\citep{simonyan2014very} is a CNN architecture known for its uniform design and depth. It features 16 layers, including 13 convolutional layers and three fully connected layers. VGG16's simplicity and depth allow it to capture complex image patterns effectively. In architectural designs like VGG16, gradients (which indicate the direction and magnitude of adjustments to network weights during training) become very small during backpropagation in deep networks. This can hinder the training of deep networks because small gradients lead to very slow learning or no learning. This phenomenon is called the vanishing gradient. 

ResNet~\citep{he2016deep} is a CNN architecture known for effectively training deep networks. ResNet50 is a variant of ResNet with 50 layers. The key innovation in ResNet is the skip connection, which mitigates the vanishing gradient problem and enables the training of much deeper networks. Skip connections skip some layers of the ResNet50 network to facilitate the flow of gradients during backpropagation and make the network easier to train. Skip connection is expressed as \(h = F(x) + x\), where \(x\) is an input, \(F(x)\) is the learned feature maps from the residual function, and \(h\) denotes the output. ResNet50 has four blocks, each a group of convolutional layers that work together to extract features. These blocks are responsible for different feature extraction levels, with the features' complexity increasing from the first to the last layer. There is a fully connected layer at the last. Wide ResNet50~\citep{zagoruyko2016wide}, a variant of wide ResNet, is an extension of the ResNet50 architecture that focuses on increasing convolutional layers' width (number of channels). 

Inception v3~\citep{szegedy2017inception} represents a modular CNN design. 
InceptionNet's modular design uses parallel convolutional paths with varying filter sizes in Inception modules to enhance feature extraction efficiency and performance in visual recognition tasks. Adopting parallel convolutional paths reduces the computational complexity. 

Deeper models tend to generate more parameters, while multiple convolutional operations decrease the size of feature maps, thereby reducing resolution. Networks often adjust filter widths arbitrarily, resulting in an uneven depth, width, and resolution distribution. This lack of uniformity has posed challenges in industrial applications, contributing to heightened costs and limited available resources~\citep{tan2019efficientnet}. EfficientNet addresses the challenge of balancing depth, width, and resolution in CNNs to improve accuracy and efficiency. Unlike previous models that randomly scaled these dimensions, It uses a compound scaling method, expressed mathematically as \( \text{Depth}(d) = \alpha^\phi \), \( \text{Width}(w) = \beta^\phi \), and \( \text{Resolution}(r) = \gamma^\phi \), with constants \( \alpha, \beta, \) and \( \gamma \) optimized through grid search. \(\alpha\) scales the depth, \(\beta\) scales the width, \(\gamma\) scales the resolution of input images, and \(\phi\) uniformly scales all network dimensions. There are different variants of EfficientNet. Among them, EfficientNet-B7 is the best-performing network in the ImageNet dataset. 

The growth of mobile devices encouraged researchers to build CNNs suitable for mobile and embedded devices with computational and memory limitations. This network is called MobileNet~\citep{howard2017mobilenets}. The main innovation in the MobileNet network is using standard convolution operations with depthwise separable convolutions (DSCs) that minimize model parameters. DSCs perform a $3 \time 3$ convolution with a single channel, sliding over the input tensor to produce an output channel for each convolution. Following this, a $1 \time 1$ pointwise convolution is used to adjust the channel depth, drastically reducing computational complexity and the number of parameters compared to conventional convolutions. MobileNet-v3~\citep{howard2019searching} is the advanced version of the MobileNet series, designed to optimize both accuracy and efficiency for mobile and embedded devices. 

\subsection{Transformer}
In 2017, a network called transformer~\citep{vaswani2017attention} came up with working in a sequence of input data by capturing long-range dependencies between the input sequence. This ability is facilitated by splitting the long sequence into small tokens and providing positional information for each token. The mechanism of self-attention is used to extract the features, where each position in the input sequence can attend to all other positions, allowing the model to relate elements that are distant from each other in the sequence. It parallelizes the processing of every pair of tokens, enabling direct interaction between all pairs. Importantly, this uniform processing ensures that each pair of tokens is treated in the same consistent manner, enhancing the model's ability to capture complex relationships across the entire sequence efficiently. Unlike CNNs, which have limited receptive fields~\citep{naseer2021intriguing}, transformers can capture information from the entire sequence due to their self-attention mechanism~\citep{zimerman2023long}. This ability to incorporate global features is particularly advantageous in tasks where understanding the entirety of the input sequence is crucial, such as natural language understanding and image classification. This subsection describes the foundational transformer networks used in object recognition.

The Vision Transformer (ViT)~\citep{dosovitskiy2020image} is an innovative approach to image classification that leverages the transformer architecture. Unlike CNNs that rely on convolutions to capture spatial hierarchies, ViT divides an image into a sequence of fixed-size overlapping patches and treats each patch as a token. These tokens are then fed into a standard transformer encoder, allowing the model to capture global features and complex relationships across the entire image~\citep{touvron2022deit}. ViT-Base is the network with a stack of 12 transformer encoders that extracts meaningful features using a multi-head self-attention mechanism. The last two layers of the ViT are layer normalization and classification token layer. The layer normalization helps to stabilize the training but adds more computational overload. Moreover, the classification token is the abstract information aggregating the representation of the entire image for classification purposes. 

Swin transformer~\citep{liu2021swin} is an improvement of the ViT. Unlike ViT, Swin introduces a hierarchical structure where images are initially divided into non-overlapping patches and then grouped into hierarchical blocks. This approach allows the model to capture both local details and global context effectively. Swin Transformer also employs shifted windows within each block, enabling overlapping receptive fields across layers to better capture diverse features at different scales. Swin transformer's tokenization strategy further enhances its ability to handle images of varying resolutions and complexities, making it suitable for tasks requiring detailed spatial information and context across multiple scales. 

\section{Related Work}
Feature fusion is a significant part of deep learning, which combines information from multiple layers or sources. It enhances the network's ability to understand and interpret complex patterns in data. The addition of different layers' features is common in deep networks like ResNet~\citep{he2016deep}, Wide ResNet~\citep{zagoruyko2016wide}, ViT~\citep{dosovitskiy2020image}, and FPN~\citep{lin2017feature}. Similarly, the concatenation of different layers of the features is present in popular networks like InceptionNets~\citep{szegedy2017inception}, DenseNet~\citep{huang2017densely}, and U-Net~\citep{ronneberger2015u}. Methods such as addition and concatenation are considered context-unaware feature fusion techniques. They treat all features equally without distinguishing between more informative and less useful ones. These methods simply aggregate features by either summing them element-wise (in the case of addition) or placing them side by side (in the case of concatenation) without any mechanism to evaluate the importance of each feature. As a result, they can include noise and irrelevant information in the final feature representation, potentially degrading the network's performance~\citep{iandola2016squeezenet, woo2018cbam, dai2021attentional}. There are some techniques like SENet (Squeeze-and-Excitation Network)~\citep{iandola2016squeezenet} and CBAM (Convolutional Block Attention Module)~\citep{woo2018cbam} to enhance the feature representations by selecting more informative features. SENet and CBAM use attention mechanisms to enhance feature fusion by focusing on the most relevant information. SENet recalibrates channel-wise feature responses by explicitly modeling interdependencies between channels. CBAM sequentially applies channel and spatial attention to emphasize informative features across both dimensions. These attention mechanisms allow the networks to selectively highlight useful features and suppress less useful ones, resulting in improved performance. However, these techniques are useful in fusing features of the same networks because these attention techniques focus on refining the features but cannot fuse features of CNN and transformer directly. This study focuses on improving cattle identification by combining features from CNN and transformer. 

This section first explores various research based on the fusion of CNN and transformer features in different domains. Based on the fusion method, they are categorized into three methods (addition-based, concatenating-based, and attention-based). Moreover, this section reviews existing research on muzzle-based cattle identification using deep-learning networks. 

\subsection{Addition Based Fusion}
Addition-based feature fusion is a simple addition of two different features. This part covers the feature fusion of two types of features with the same or different scales and resolutions applied in various areas. 

\citet{nie2023cross} proposed simply adding the features of two different networks with the same scale. It introduced a cross-modal feature fusion strategy using CNNs and transformers, optimizing feature diversity and intermodal information exchange through the Cross-Modal feature fusion. \citet{song2022ctmfnet} extracted multi-scale local features using CNN and global features from the transformer; both features were passed through the convolution layer to match the channel dimension and finally added. The fused feature tensor was further refined, enhancing detailed features within target regions and suppressing irrelevant information from surrounding areas. This enriched the overall feature representation, improving segmentation and detection tasks.
Similarly, multi-scale features were extracted and added in the research~\citep{chen2023hybrid}. It addressed the challenges of effectively removing rain streaks from single images exhibiting complex geometric appearances and overlapping phenomena. Traditional methods struggle due to the diverse nature of rain patterns and their irregularities. The study proposed that the hybrid CNN-transformer network could overcome these limitations by integrating them progressively. In the CNN-based stage, spatially varying rain distribution features were present. In the transformer-based stage, background-aware features were extracted by capturing long-range feature dependencies for global texture recovery while maintaining structural integrity. CNN and transformer features were recalibrated using channel attention network SENet~\citep{hu2018squeeze} and added with original CNN and transformer features. Multi-scale and multi-resolution features were added in the research~\citep{wang2022net}. It addressed the limitation of U-Net in medical images as U-Net struggles to capture global and long-range dependencies. It used a CNN and transformer dual branch, aggregating the local and global extracted features. In addition, features from a CNN and a transformer having different channel dimensions were passed through the $1 \times 1$ convolution layer to align the dimension and finally added~\citep{peng2021conformer}. \citet{zhu2024diffusion} used weighted summation of the different features. The weighted summation allows more control over which features contribute more significantly to the final aggregated feature, enabling a more refined and potentially more effective representation. \citet{vindas2022hybrid} used a dual branch, multi-scale feature extraction using transformer and CNN, followed by the weighted addition of the features. The weighted addition is better than a simple addition as it uses Hadamard product~\citep{horn1990hadamard} to include more informative features. \citet{yang2023mmvit} proposed a method to weigh the importance of different features by calculating attention score. It generates a score map representing the quality scores of different patches in an input image. It also generates an attention map representing a matrix that assigns weights or importance to each patch in the input image. The attention score is obtained through a weighted summation of the score map and attention map using the Hadamard product. This mechanism dynamically adjusts the contribution of each patch based on its significance, simulating the human visual system's focus on essential areas. The Hadamard product performs simple element-wise interactions and cannot capture complex dependencies and context within a sequence to simultaneously focus on different input parts.

\subsection{Concatenation Based Fusion}
Concatenation of features refers to combining multiple feature vectors into a single feature vector. This is a common technique in machine learning and data processing to create a more comprehensive representation of features. Features with different scales and resolutions are appended or stacked together using concatenation. This subsection covers some notable research used for feature fusion.

Multi-scale features were concatenated by the research~\citep{Jiang_2023} to detect objects of varying sizes and complexities. It integrated CNNs with a transformer and deformable convolutions in the feature extraction network. This fusion approach enhanced feature extraction capabilities by leveraging ViT's ability to capture global dependencies alongside CNN's proficiency in spatial feature extraction. This approach was applied to accurately detect marine organisms of varying sizes and complexities. \citet{Wei_2022} tackled the challenges in semantic segmentation of remote sensing images, where traditional methods struggled to utilize abundant semantic information and irregular shape patterns effectively. Convolutions and single-scale feature maps often fail to capture the diverse contextual details required for accurate segmentation. A multi-scale feature pyramid detector was proposed~\citep{Wei_2022} to fuse image features. The proposed decoder employed the 2D-to-3D transform methods to obtain multi-scale feature maps that contained rich context information and fuse the multiscale feature maps channel using concatenation. Similarly, \citet{yu2021unsupervised, xu2024fcsu} concatenated CNN and ViT features simply to provide better accuracy and generalization to the new data.

\citet{dutta2023conv} concatenated three types of features with different scales and resolutions. This research emphasized CNNs for local pixel correlations and a vision transformer for capturing long-range pixel correlations to extract shape-based features. Similarly, multi-scale and multi-resolution features were concatenated in medical images in the paper~\citep{Oukdach_2023}. It addressed the challenges of using ViT in medical imaging, particularly with small-size datasets such as those in wireless capsule endoscopy. \citet{nguyen2023collaborative} extracted CNN and ViT features from the medical X-ray images and concatenated them. \citet{yang2023hyperspectral} proposed an innovative approach to fuse CNN and transformer features. Both features from different depths were fused along the channels and refined using convolutional blocks. \citet{wang2022vit} addressed the challenges of semantic segmentation, characterized by abundant semantic information and irregular shape patterns. It used transformer features and applied CNN to provide spatial and channel attention, followed by channel-wise concatenation to features. \citet{oukdach2023conv} presented a robust approach to enhancing ViT performance for medical image classification with limited data. By integrating CNN modules to extract detailed features and concatenating them with ViT's global representations, the proposed model achieved notable improvements in accuracy and robustness. \citet{Qiao_2023} integrated a Feature Pyramid Network (FPN)~\citep{lin2017feature} to enhance cooperative perception among connected autonomous vehicles. The FPN extracted multi-resolution and multi-scale intermediate features through three downsample blocks from the pseudo-image representations derived from point cloud data. These intermediate features were concatenated across the channel and spatial dimensions. This fusion mechanism consolidated information from various viewpoints and resolutions, enriching the feature representations crucial for accurate perception in dynamic urban environments. \citet{lee2023plant,li2024vtcnet} extracted multi-scale and multi-resolution features from CNN and ViT networks by concatenating them to develop a better recognition method. 

\subsection{Attention Based Fusion}
Attention is how a certain portion of the input image is provided with more focus. The attention-based methods~\citep{Dai_2020, Ding_2023, Zhou_2022} were used for certain parts of the features. However, each attention fusion used different pre- and post-processing methods to filter the best features. Moreover, there are research~\citep{carion2020end, ma2021facial, wang2022hybrid, qingyun2021cross, chen2023thfuse, zeng2022nlfftnet, xia2024vit, xu2024joint} that provided an attention mechanism to the features extracted by CNN by feeding the CNN features into the transformer networks. This type of CNN and transformer fusion does not directly aggregate the features. These are series-based integrations of CNNs into the transformer but not direct feature fusion.

\citet{qi2022ftc, xing2024multi} extracted different features, provided channel-wise attention to both features, and combined them to obtain the attention mask using a softmax. The attention mask was multiplied by the transformer features. This step adjusted the transformer features by emphasizing the parts deemed important by the attention mask and de-emphasizing the less important parts. \citet{zhou2024research} applied CBAM on local and global features to select the most informative features and fused. 

\citet{chen2022mobile} proposed a two-way bridge in the Mobile-Former architecture that facilitates communication between the local features of MobileNets and the global information of transformers. In the ``Mobile to Former" direction, local features were sent to the transformer part using lightweight cross-attention, efficiently merging them with global tokens without complex computations. Conversely, in the ``Former to Mobile" direction, global tokens enhanced the local features when sent back to the MobileNets part, providing a broader context. Integrating detailed local features and global context improved the architecture's final performance in image classification tasks. But, it struggled with speed and accuracy on smaller images. 

\subsection{Existing Cattle Identification Methods}
Deep learning models, particularly Convolutional Neural Networks (CNNs), have been extensively applied in cattle identification research. Among the most utilized architectures are VGG, ResNet, Wide ResNet, InceptionNet, EfficientNet, and MobileNet, reflecting the effectiveness of CNN-based models in extracting discriminative features from cattle images.

VGG16 has demonstrated superior performance across multiple studies, often surpassing other CNN variants in cattle identification tasks. VGG16 outperformed the different CNNs reported by the research~\citep{li2022generalized, guzhva2018now, wu2021using, li2019deep, wang2020cattle, rivas2018detection}. This is likely due to its deep architecture and consistent hierarchical feature representation, making it well-suited for image-based classification tasks. ResNet was utilized by the research~\citep{shojaeipour2021automated, bezen2020computer,salau2020instance, liu2020video, xu2020automated, qiao2019cattle, xiao2022cow} showcasing the strength of CNN in identifying cattle effectively. \citet{kimani2023cattle} utilized Wide ResNet as an effective cattle identification method. Wide ResNet leverages greater capacity regarding the number of features learned per layer. Inception Net demonstrated superior performance reported by the research~\citep{qiao2019individual, qiao2021automated, ren2021tracking, porter2021feasibility}. EfficientNet was used by the research~\citep{lee2023identification, saar2022machine, yin2020using} for the optimum performance of identifying Hanwoo cattle. Moreover, lighter networks like MobileNets were utilized to develop a lightweight cattle identification by the research~\citep{li2021individual, hou2021cow, zin2020automatic, xudong2020automatic}. ViT was also applied to explore the promising alternative to traditional CNN-based approaches by some cattle identification research like~\citep{bergman2024biometric, wu2021body, guo2023vision, zhang2023high}. Additionally, the Swin transformer was explored to expand the application of transformer models in cattle identification by some research like~\citep{lu2022recognition, zhao2023research, zhong2023method}. 

\subsection{Feature Fusion in Cattle Identification}
Deep feature fusion has been applied in a few research studies in cattle identification. \citet{ojala1994performance} extracted features by using CNNs followed by addition to improve cattle identification. A combination of CNN and transformer has benefited the face recognition of sheep cattle~\citep{li2023combining}. However, this research~\citep{li2023combining} used simple convolution networks to extract the features and feed them into a transformer to extract global features. Finally, the features extracted by CNN and the transformer were fused by addition. \citet{du2022individual} extracted local features from the earlier layers of VGG16 and global features from the higher layers of the VGG16 and fused by concatenating local and global features. \citet{wan2023sheep} used VGG16 to extract features in two branches, one with features with spatial attention and the other with channel attention, and finally concatenated. Ghost convolution~\citep{han2020ghostnet} and CBAM were used by the research~\citep{fu2022lightweight} in ResNet50 to provide channel and spatial attention in the feature extraction process. \citet{hu2020cow} used different CNN features from different parts of the cattle, and then the features were added to enhance the identification accuracy. \citet{weng2022cattle} used dual-branch CNNs to extract features, and the extracted features were channel-wise recalibrated using SE block followed by a concatenation of both branches' features. 

Most existing research in cattle identification has focused on using either CNNs or transformers, rarely integrating both. Even in cases where feature fusion is employed, the approach is typically limited to direct addition or simple concatenation of features. In this work, a novel dual-branch method for cattle identification is proposed. One branch extracts local features using CNN, and another extracts global features using a transformer. Both branches work parallel, and both features are fused using a multi-head attention mechanism. Multi-head attention considers the long-range relationship between CNN and the transformer features. The most informative features are selected based on the attention score.

\section{Methodology}
We took a representative ResNet50 as CNN and ViT as a transformer network and modified both to reduce the computational complexity. ResNet-50 was chosen for its balance of depth, simplicity, and proven reliability in various image recognition tasks. Its skip connections effectively mitigate the degradation problem in deep networks, allowing for more efficient training of complex networks~\citep{prajwal2023comparative, suherman2023implementation}. ResNet50's widespread adoption and availability of pre-trained models make it a practical choice for many image recognition applications~\citep{durga2023convolutional, shojaeipour2021automated, zhang2023transfer, hong2022resdnet}. Similarly, ViT was chosen for its excellent performance in image recognition in ImageNet data, leveraging self-attention mechanisms to capture long-range dependencies in images more effectively than traditional CNNs~\citep{dosovitskiy2020image}. 

\begin{figure}[ht]
    \centering
    \includegraphics[width=1\linewidth]{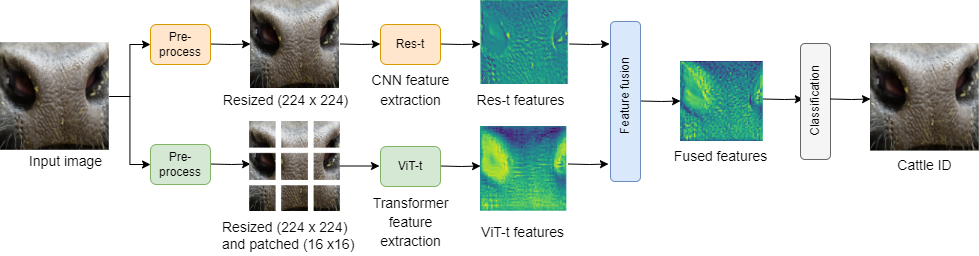}
    \caption{Architecture of the dual branch MHAFF method for cattle identification. The CNN branch extracts local features, while the ViT branch captures global features. These features are fused using a multi-head attention mechanism with a context-aware approach, enhancing identification performance}
    \label{fig:fig1}
\end{figure}

The sketch of the proposed method is shown in Fig.~\ref{fig:fig1}. The model processes input images through a CNN and a transformer to extract distinct features. These extracted features are then combined using multi-head attention mechanisms. After fusion, the combined features are processed through a fully connected layer and classified using a softmax layer. 

\subsection{Datasets Description}
In this study, we used two benchmark datasets and two publicly available cattle datasets. The benchmark datasets used were the CIFAR10~\citep{krizhevsky2009learning} and Flower102~\citep{nilsback2008automated}. 
CIFAR10 (Canadian Institute for Advanced Research - 10) is a widely used image classification dataset. It consists of 60000 images, each with dimensions of $32 \times32$ pixels, distributed across ten different classes. This dataset is split into 50000 training and 10000 testing images.
Flower102, also known as the Oxford 102 Flowers dataset, is an image dataset designed for fine-grained visual categorization. It comprises 102 different classes of flowers. The images vary in scale, pose, and lighting conditions, providing a challenging dataset for training and evaluating fine-grained recognition models. The total number of images is 7169, with 1020 training and 6149 testing images. The image sizes range from $500 \times 500$ pixels to $500 \times 1168$ pixels, providing diverse visual representations for model training and evaluation. 

\begin{figure}[ht]
    \centering
    \includegraphics[width=0.5\linewidth]{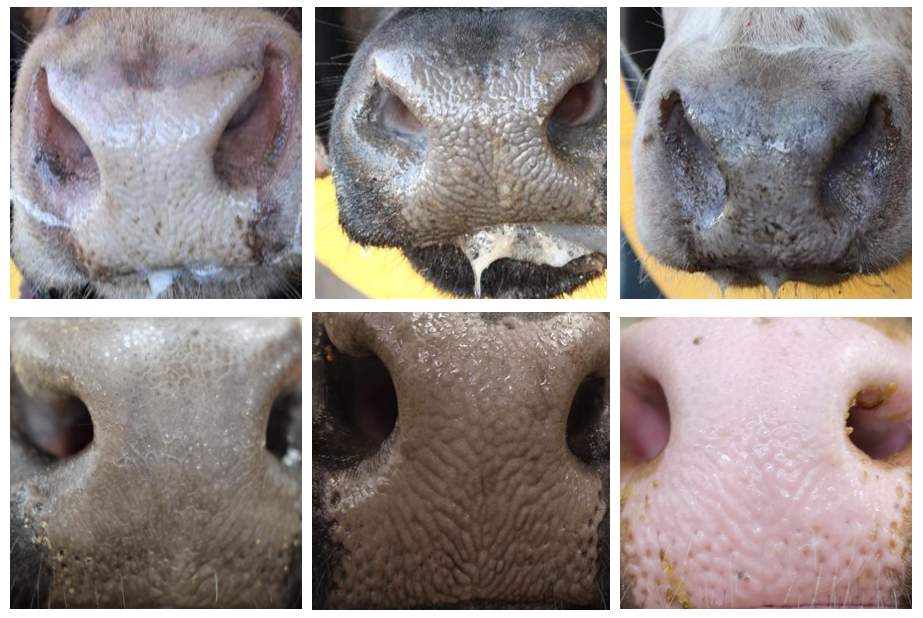}
    \caption{Sample of the muzzle images of Cattle-1 and Cattle-2 datasets. The upper row represents samples for Cattle-1, and the lower row represents samples for Cattle-2}
    \label{muzzle}
\end{figure}

We used two publicly available cattle datasets: Cattle-1~\citep{shojaeipour2021automated} and Cattle-2~\citep{li2022individual}. Cattle-1 data was captured by researchers at the University of New England (UNE), Australia, on the university farm. This data includes 2632 cattle face images of 300 cattle captured on a sunny day under natural light conditions from the 1-2 meter front of the cattle face and 1 meter above the ground. The muzzles are detected and extracted from the facial image data using a modified YOLOv5 model developed from our previous research~\citep{dulal2022automatic}. The extracted muzzle dataset has a total of 2447 images. From the 2447 images, the dataset is randomly split into training, validation, and testing in a proportion of 70\%, 20\%, and 10\%, respectively. The lowest resolution is $200$ (width) $\times$ 400 (height) pixels, and the highest is $600 \times 600$. This data contains images from different breeds, including Angus, Hereford Charolais, and Simmental. The images are full-faced with the cattle in various colors, such as white, black, brown, and red. The number of images per cattle is between 6 and 16. The first row of three images shown in Fig.~\ref{muzzle} are sample images for the Cattle-1 dataset. 

Cattle-2~\citep{li2022individual} data was captured by the researchers at the University of Nebraska-Lincoln (UNL) Eastern Nebraska Research Extension and Education Center (ENREEC)'s farm located in the United States of America (USA). This data includes 4923 images from 268 mixed breed cattle (Angus, Angus- Hereford cross, and Continental- British cross), captured from outside the pen from the front side with varying distances. In contrast, cattle were inside the pen in a natural lighting condition. The available images are muzzle regions obtained from manual cropping. The range of the images per cattle is between 4 and 70. The lowest resolution is $68 \times 44$, and the highest is $3104 \times 2704$. Sample images for the Cattle-2 dataset are shown in the second row in Fig.~\ref{muzzle}. This data was also split into training, validation, and testing at the proportion of 70\%, 20\%, and 10\%, respectively. 

\subsection{Model Setup}
The MHAFF method involves two different types of networks for feature extraction. Both feature extraction networks take different types of image inputs. The transformer sequentially processes images, treating input images into sequences of patches and the position of each patch. In contrast, CNNs don't inherently consider the sequence of the image. To ensure compatibility with both architectures, we prepared the data accordingly. CNN transforms a full image using Torchvision's library called transforms. This transformation involves resizing the input image to a dimension of 256, followed by a central cropping to extract a $224 \times 224$ region. The image is then converted to a PyTorch tensor and normalized. These setups ensure the input data is appropriately formatted and standardized before being fed into the CNNs during training. On the other hand, the transformer uses sequences of patches of the images. Image transformation for the transformer involves resizing all images to a fixed size of $224 \times 224$ pixels, and the image is divided into patches of $16 \times 16$. This transformation standardizes the input images, ensuring consistent dimensions and pixel value ranges as part of the data preparation for the transformer model. 

Moreover, data augmentation techniques were employed to boost the number and variety of images during training. This approach helps to synthetically generate new images and expand small datasets for training deep learning models. In this study, data augmentation was applied during the preprocessing stage. Four different strategies were utilized: horizontal flipping, brightness modification, random rotation, and blurring. Horizontal flipping was introduced to mimic different positions of cattle captured due to their natural movement. Brightness modification simulated various outdoor lighting conditions, with brightness values adjusted between 0.2 and 0.5 on a scale where 0 is the darkest and 1.0 is the brightest. Random rotation, set between -15° and 15°, was used to replicate the natural head movements of cattle. To account for overexposure and motion blur, blurring was applied using a Gaussian filter with kernel sizes ranging from 1 to 5, resulting in blurred muzzle images.

\subsection{Feature Extraction}
Feature extraction is a process where raw data (such as images, signals, or text) is transformed into a set of measurable and meaningful attributes called features~\citep{khalid2014survey, mutlag2020feature}. These features can be used for analysis and model training. The purpose of feature extraction is to reduce the complexity of the data while retaining essential information that captures the underlying patterns relevant to the specific task, such as classification, detection, or recognition~\citep{liu1998feature, khalid2014survey, mutlag2020feature}.

Modified ResNet and ViT networks are used to extract features. The lower layers of ResNet50 can extract local features~\citep{chen2022holstein}. However, ViT extracts mixed features in the lower layers, but the higher layers extract global features~\citep{dosovitskiy2020image}. The feature representations of ResNet50 and ViT networks reveal that the lower half of the ResNet50 and the lowest quarter of the ViT are highly similar~\citep{raghu2021vision}. Thus, ResNet50 and ViT are modified to extract the local and global features accordingly. Res-t is a modified version of ResNet50 that removes the last dense layer and fine-tunes the lower half portion. The lower half layers mean the first two blocks of the ResNet50. Similarly, 
ViT-t is the modified version of ViT that removes the last two layers and fine-tunes the upper eight encoder layers. 

\subsection{Feature Fusion}
Feature fusion combines multiple sets of features from different sources or feature extraction methods into a single, unified representation. Feature fusion aims to leverage complementary information from different feature sets to improve the performance of a deep learning network. This section explains the proposed feature fusion method based on multi-head attention. 
Multi-head attention provides information on where to pay attention in the feature maps. It provides more focus across spatial domains and uses it
to select important spatial regions~\citep{hu2018gather, wang2018non} or find
the most relevant spatial position directly~\citep{dai2017deformable, mnih2014recurrent}. Multi-head attention is also used to gather spatial attention between the Res-t and ViT-t features~\citep{guo2022attention}. MHA uses three important parameters called Query (Q), Key (K), and Value(V). Q is the entity that tries to compute new representations. K and V are the pairs that pay attention based on the Q. The fusion of CNN and transformer features using MHA pays attention to both features for the given representations. In multi-head self-attention~\citep{vaswani2017attention}, if $S$ represents the input sequence and Q, K, and V are calculated by multiplying with learnable weight matrices, using the formula:
\begin{equation}
\label{eq:qkv1}
Q = S W^{Q}, \quad K= S W^{K}, \quad V = S W^{V}
\end{equation}
where \( W^{Q} \), \( W^{K} \), and \( W^{V} \) are learnable weight matrices, which are assigned randomly at first and are updated during training.

\begin{figure}
    \centering
    \includegraphics[width=0.4\linewidth]{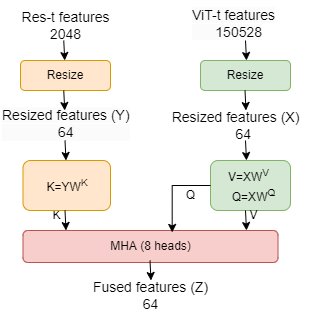}
    \caption{Proposed multi-head attention method. This clearly shows the generation of Q, K, and V in a different manner than the original transformer}
    \label{att}
\end{figure}

The proposed method generates the query (Q), key (K), and value (V) matrices using feature vectors generated from Res-t and ViT-t. Specifically, as illustrated in Fig.~\ref{att}, the key (K) matrix is derived from the features obtained by Res-t, while the query (Q) and value (V) matrices are derived from the features obtained by the ViT-t. Q, K, and V dimensions are the same, represented by $d_q$, $d_k$, and $d_v$. Let $X$ represent a matrix of transformer features from ViT-t, and $Y$ represent a matrix of CNN features from Res-t. We tried different combinations of $X$ and $Y$ to generate $Q$, $K$, and $V$. There are six possible combinations to generate $Q$, $K$, and $V$ from $X$ and $Y$. We ran the experiments on all possible combinations, and the highest accuracy was obtained on ($X$,$Y$,$X$). Table~\ref{res:qkv} shows the accuracy of each combination to generate Q, K, and V. X and Y represent features from ViT-t and Res-t, respectively.  

Q, K, and V are generated by using the formula:

\begin{equation}
\label{eq:qkv}
Q = X W^{Q}, \quad K= Y W^{K}, \quad V = X W^{V}
\end{equation}

Attention is calculated by using a scaled dot-product between Q and K, applying the softmax function to the result, and then multiplying by V. The scaling factor, $\frac{1}{\sqrt{d_k}}$ is used to scale the dot product between Q and K. 
The attention mechanism can be expressed as:

\begin{equation}
\text{Attention}(Q, K, V) = \text{softmax}\left(\frac{Q K^T}{\sqrt{d_k}}\right) V
\end{equation}

The input sequence is divided into different numbers called heads to process the attention in parallel. Attention is calculated across different heads, and the final attention is calculated by concatenating the results of each parallel head. For multi-head attention (MHA) with $h$ heads, the process can be generalized as:

\begin{equation}
\text{MHA}(Q, K, V) = \text{Concat}(\text{head}_1, \text{head}_2, \ldots, \text{head}_h) W^O
\end{equation}
 \(W^O\) is a learnable parameter assigned randomly and updated during training. The fused feature \( Z \) is then calculated using the MHA process:

\begin{equation}
Z = f(\text{MHA})
\end{equation} 

The fused feature vector Z, derived from the multi-head attention (MHA) mechanism, has a dimensionality of 64. The vector is subsequently passed through a fully connected layer to produce an output vector \( \mathbf{Z}' \) of dimension \( C \), corresponding to the number of cattle classes. This transformation is mathematically expressed as:
\begin{equation}
 \mathbf{Z}' = \mathbf{A} \mathbf{Z} + \mathbf{b} 
 \end{equation}
where \( \mathbf{A} \) denotes the \( C \times 64 \) weight matrix and \( \mathbf{b} \) represents the bias vector of dimension \( C \).

Following the transformation, a softmax activation function is applied to the elements of \( \mathbf{Z}' \) to compute the probabilities \( P(\text{cattle}=k \mid \mathbf{Z}') \) for each class \( k \). The softmax function normalizes the output scores into a probability distribution, defined as:
\begin{equation}
 P(\text{cow}=k \mid \mathbf{Z}') = \frac{e^{Z'_k}}{\sum_{l=1}^{C} e^{Z'_l}}
\end{equation}
where \( Z'_k \) denotes the \( k \)-th element of \( \mathbf{Z}' \).

This approach allows the classification of cattle within a categorical framework. The model outputs class probabilities via softmax activation by leveraging the transformed feature vector \( \mathbf{Z}' \). These probabilities \( P(\text{cattle}=k \mid \mathbf{Z}') \) indicate the likelihood of the fused features \( \mathbf{Z}' \) belonging to each cattle class \( k \), enabling identification of cattle based on the model's predictions.


\subsection{Model Training and Validation}
Transfer learning was applied to Res-t and ViT-t networks, leveraging pre-trained weights from the ImageNet~\citep{deng2009imagenet} dataset through PyTorch's library. Transfer learning is a powerful technique for better performance, especially when dealing with small datasets using transformers. It allows for faster training, better generalization, and efficient use of limited data and computational resources~\citep{liu2021efficient}. The experiments were conducted with 50 epochs of training. Pre-trained weights of our proposed networks were fine-tuned by minimizing multi-class cross-entropy loss. Several optimization approaches, such as dynamic learning rate adjustment and early stopping, were also applied to avoid overfitting. The Adam optimizer and the ReduceLROnPlateau technique for dynamic learning rate adjustment were used during training. The learning rate was reduced by 0.1 if the loss didn't improve for five epochs, with an initial learning rate set to \(1 \times 10^{-5}\). Early stopping was employed with a patience of 20 epochs and a dropout rate of 0.3. All hyperparameters relevant to transfer learning, such as epochs, learning rate, early stopping, and dropout, underwent meticulous adjustment across numerous training iterations.

The loss function ($\lambda_{\text{MHAFF}}$) is defined as follows:

\begin{equation}
\lambda_{\text{MHAFF}}(o, P) = -\frac{1}{N} \sum_{i=1}^{N} \sum_{k=1}^{C} o_{i,k} \cdot \log(P_{i,k})
\end{equation}
where $N$ is the batch size, $C$ is the number of classes, $o_{i,k}$ is an indicator (0 or 1) of whether class $k$ is the correct identification for sample $i$, and $p_{i,k}$ is the predicted probability of sample $i$ belonging to class $k$.

The accuracy metric quantifies the proposed method's predictive performance in the validation phase. This accuracy is ascertained by comparing the model's predictions, obtained through forward propagation, against the veritable labels. The validation accuracy, expressed as the ratio of correctly predicted samples to the total samples in the validation dataset, is calculated as follows:

\begin{equation}
\text{Validation Accuracy} = \frac{\text{Number of Correct Predictions}}{\text{Total Number of Validation Samples}}
\end{equation}

\section{Results and Discussion}

\subsection{Selection of Q, K, and V}
We first used the Flower102 and CIFAR10 datasets to determine the best Q, K, and V combination. The best combination identified from these datasets was subsequently applied to the cattle datasets. The Flower102 and CIFAR10 datasets were chosen due to their status as small-sized benchmark datasets. This facilitates efficient and reliable evaluation of Q, K, and V combinations to generate the highest attention score. Table~\ref{res:qkv} presents the results of the different combinations of Q and V values from ViT-t feature inputs and K values from Res-t feature inputs in both datasets. The proposed combination (Q=X, K=Y, and V=X) outperformed the second-highest combination (Y, X, X) by 3.84\% and 5.21\% on Flower102 and CIFAR10, respectively. These results indicated that this combination of feature inputs for the Query, Key, and Value matrices was optimal for our task. Therefore, we selected this configuration as the best approach for our experiment, ensuring the most effective use of the model’s capabilities for accurate and robust performance.
\begin{table}[ht]
    \centering
    \caption{Results on different combinations of Q, K, and V}
    \begin{tabular}{lcc}
    \hline
        \multirow{2}{*}{Q, K, V} & \multicolumn{2}{c}{Accuracy (\%)}\\ \cline{2-3}
        &Flower102 &CIFAR10  \\\hline\hline 
        X, Y, Y & 92.64 & 94.25\\
        Y, X, Y & 93.13 & 88.73\\
        Y, Y, X & 91.87 & 91.98\\
        Y, X, X & 95.92 & 93.59\\
        X, X, Y & 95.57 & 93.06\\\hline 
       \textbf{X, Y, X} (this study) & \textbf{99.76} & \textbf{99.46}\\
\hline

\hline
    \end{tabular}
    
    \label{res:qkv}
\end{table}

\subsection{Comparison of Feature Fusion Methods}
Firstly, the performance of the proposed method is compared with that of addition and concatenation techniques. The comparative results of the addition, concatenation, and proposed feature fusion methods are presented in Table~\ref{tab:exp1}. This table also presents the results of individual feature extraction networks (Res-t and ViT-t). 
\begin{table}[ht]
    \centering
     \caption{Validation accuracy of different methods on different datasets}
    \begin{tabular}{lcccccc}
    \hline
         \multirow{2}{*}{Method} & \multicolumn{4}{c}{Accuracy (\%)}\\
         \cline{2-5}
         & Flower102 & CIFAR10  & Cattle-1 & Cattle-2 \\\hline\hline
        Res-t & 94.97 & 88.78  & 91.91 & 95.79 \\
        ViT-t & 96.06 & 89.62  & 97.62 & 95.06 \\
        Addition & 97.90 & 94.20  & 98.88 & 98.08 \\
        Concatenation & 97.26 & 95.19  & 98.36 & 97.28 \\\hline
        \textbf{MHAFF (this study)} & \textbf{99.76} & \textbf{99.46}  & \textbf{99.88} & \textbf{99.52} \\\hline

        \hline
    \end{tabular}
    \label{tab:exp1}
\end{table}
According to the results of Table~\ref{tab:exp1}, individual feature extraction networks Res-t and ViT-t perform lower than all feature fusion methods. By fusing both approaches' strengths, all combined methods (addition, concatenation, and MHAFF) can effectively leverage the local feature extraction capability of CNNs and the global context understanding of transformers. This synergy leads to improved performance. Furthermore, the MHAFF outperforms addition and concatenation in Flower102 and CIFAR10 datasets. Multi-head attention learns and captures complex interdependencies between Res-t and ViT-t features. It dynamically assigns weights to different parts of the input features based on relevance. This means that the most important features are given more focus by providing a higher attention score, resulting in a context-aware and more effective fusion. The results in Table~\ref{tab:exp1} demonstrated that our proposed method achieved superior performance on the cattle data, confirming its effectiveness. 

\begin{figure}[!t]
\centering
\begin{tabular}{>{\centering\arraybackslash}m{1.5cm} >{\centering\arraybackslash}m{2cm} >{\centering\arraybackslash}m{2cm} >{\centering\arraybackslash}m{2cm} >{\centering\arraybackslash}m{2cm} >{\centering\arraybackslash}m{2cm} >{\centering\arraybackslash}m{2cm}}
 & Original & Res-t & ViT-t & Addition & Concatenation & MHAFF \\ 

CIFAR10 & \includegraphics[width=2cm,height=2cm]{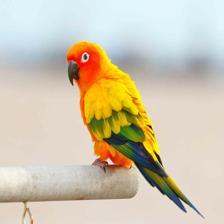} & \includegraphics[width=2cm,height=2cm]{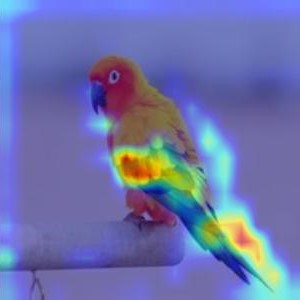} & \includegraphics[width=2cm,height=2cm]{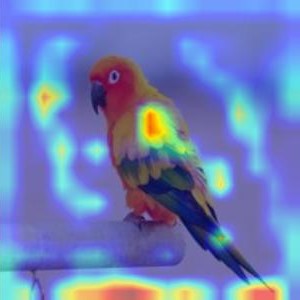} & \includegraphics[width=2cm,height=2cm]{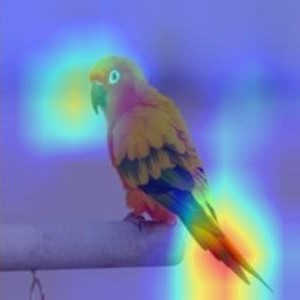} & \includegraphics[width=2cm,height=2cm]{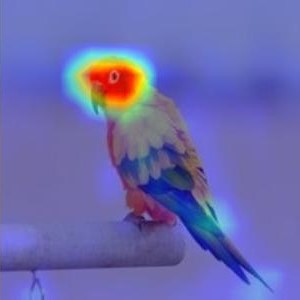} & \includegraphics[width=2cm,height=2cm]{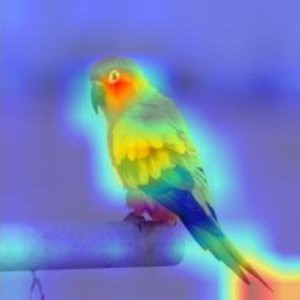} \\ 

Flower102 & \includegraphics[width=2cm,height=2cm]{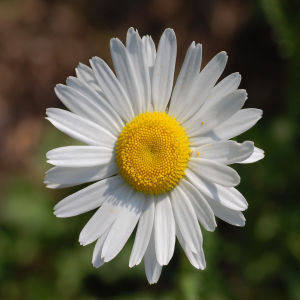} & \includegraphics[width=2cm,height=2cm]{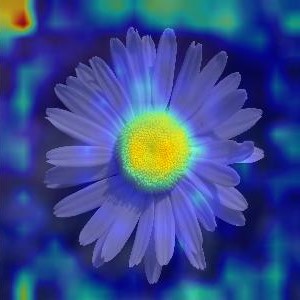} & \includegraphics[width=2cm,height=2cm]{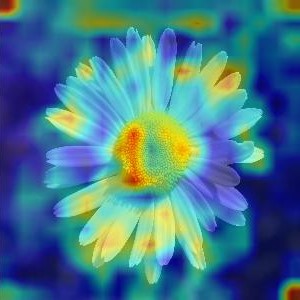} & \includegraphics[width=2cm,height=2cm]{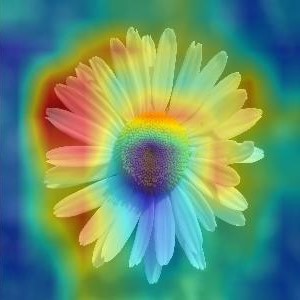} & \includegraphics[width=2cm,height=2cm]{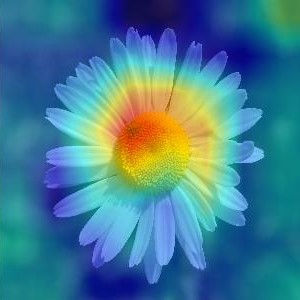} & \includegraphics[width=2cm,height=2cm]{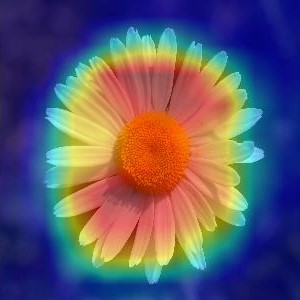} \\ 

Cattle-1 & \includegraphics[width=2cm,height=2cm]{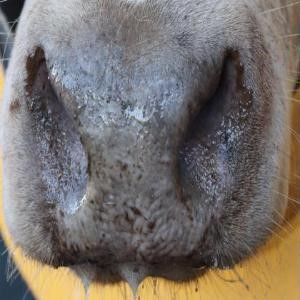} & \includegraphics[width=2cm,height=2cm]{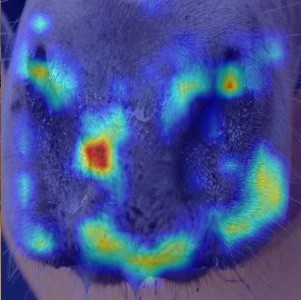} & \includegraphics[width=2cm,height=2cm]{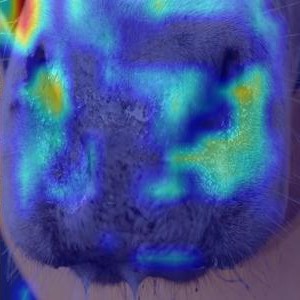} & \includegraphics[width=2cm,height=2cm]{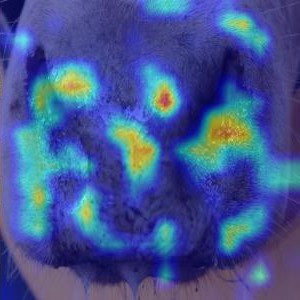} & \includegraphics[width=2cm,height=2cm]{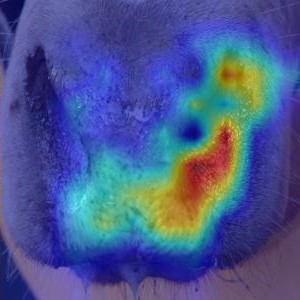} & \includegraphics[width=2cm,height=2cm]{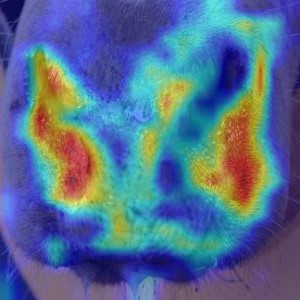} \\ 

Cattle-2 & \includegraphics[width=2cm,height=2cm]{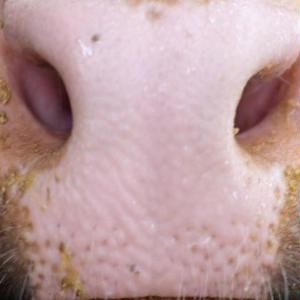} & \includegraphics[width=2cm,height=2cm]{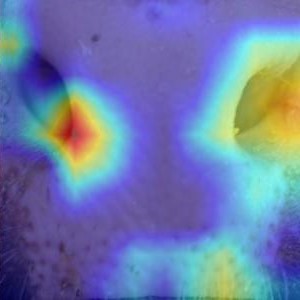} & \includegraphics[width=2cm,height=2cm]{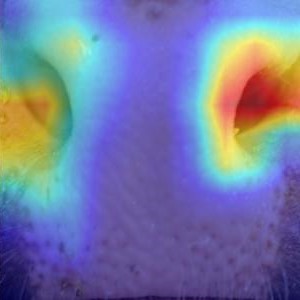} & \includegraphics[width=2cm,height=2cm]{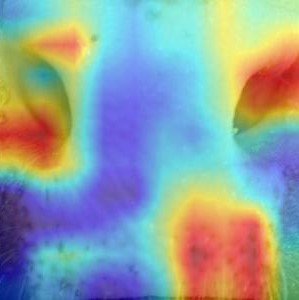} & \includegraphics[width=2cm,height=2cm]{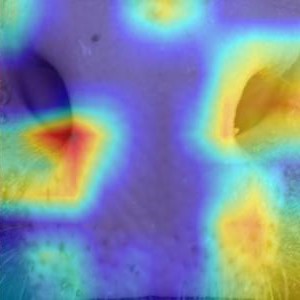} & \includegraphics[width=2cm,height=2cm]{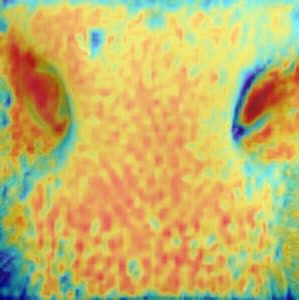} \\ 

\end{tabular}
\caption{Grad-CAM visualizations of sample images from four datasets}
\label{gradcam}
\end{figure}

We also utilized Grad-CAM~\citep{selvaraju2017grad} on the sample of images from each dataset with Res-t, ViT-t, addition, concatenation, and MHAFF methods to visualize which regions of the images contributed most to the model's predictions. This helped us interpret the internal workings of our model, verifying whether it focused on the relevant features for different classes. By examining the class activation maps, we confirmed that the model identified meaningful features, which enhanced our confidence in its generalization capabilities and performance across the dataset. It can be observed from Fig.~\ref{gradcam} that the MHAFF method consistently produced more focused and precise attention maps compared to the other methods, indicating its superior ability to extract discriminative features. 

Across all four datasets, MHAFF consistently produces the best Grad-CAM visualizations, offering the most precise attention to the relevant parts of the image. The visualizations generated by MHAFF show a clear focus on key features while minimizing the background and irrelevant parts of the image. MHAFF excels in producing more focused and context-aware activations. In particular, MHAFF generated sharper and more localized heatmaps, demonstrating its effectiveness in capturing fine-grained details. This visualization analysis reinforces the quantitative results, providing strong qualitative evidence of the robustness of our model and the advantages of the MHAFF feature fusion technique over traditional methods. The improved activation area of focus by MHAFF is because of the context-aware fusion of two features. 

\subsection{Comparison With Baseline Models}
As outlined in the related work section, the baseline models for cattle identification are VGG16, ResNet50, Wide ResNet50, Inception-V3, EfficientNet-B7, MobileNet-V3, ViT, and Swin. These models were all trained under the same experimental setup. The comparative results of these networks are presented in Table~\ref{existing}.
\begin{table}[ht]
    \centering
    \caption{Performance comparison with baseline cattle identification models}
  \begin{tabular}{lcc}
    \hline
    \multirow{2}{*}{Model} & \multicolumn{2}{c}{Accuracy (\%)}\\ \cline{2-3}
       & Cattle-1 & Cattle-2 \\\hline\hline
       VGG16\_bn  & 97.08 & 96.48 \\
        ResNet50  & 92.04 &96.50 \\
         Wide ResNet50 &96.40  &95.84  \\
         Inception-v3 & 92.48 &  93.12\\
        EfficientNet-B7  & 96.32 & 91.20 \\
        MobileNet-v3 & 97.68 & 96.80 \\
        ViT & 97.36 &  96.80\\
        Swin & 97.28 &  96.80\\\hline
        \textbf{MHAFF} (this study)  & \textbf{99.88} & \textbf{99.52} \\\hline

        \hline
    \end{tabular}
    \label{existing}
\end{table}

\begin{figure}[ht]
 \centering
 \begin{subfigure}[b]{0.45\textwidth}
     \centering
     \includegraphics[width=\textwidth]{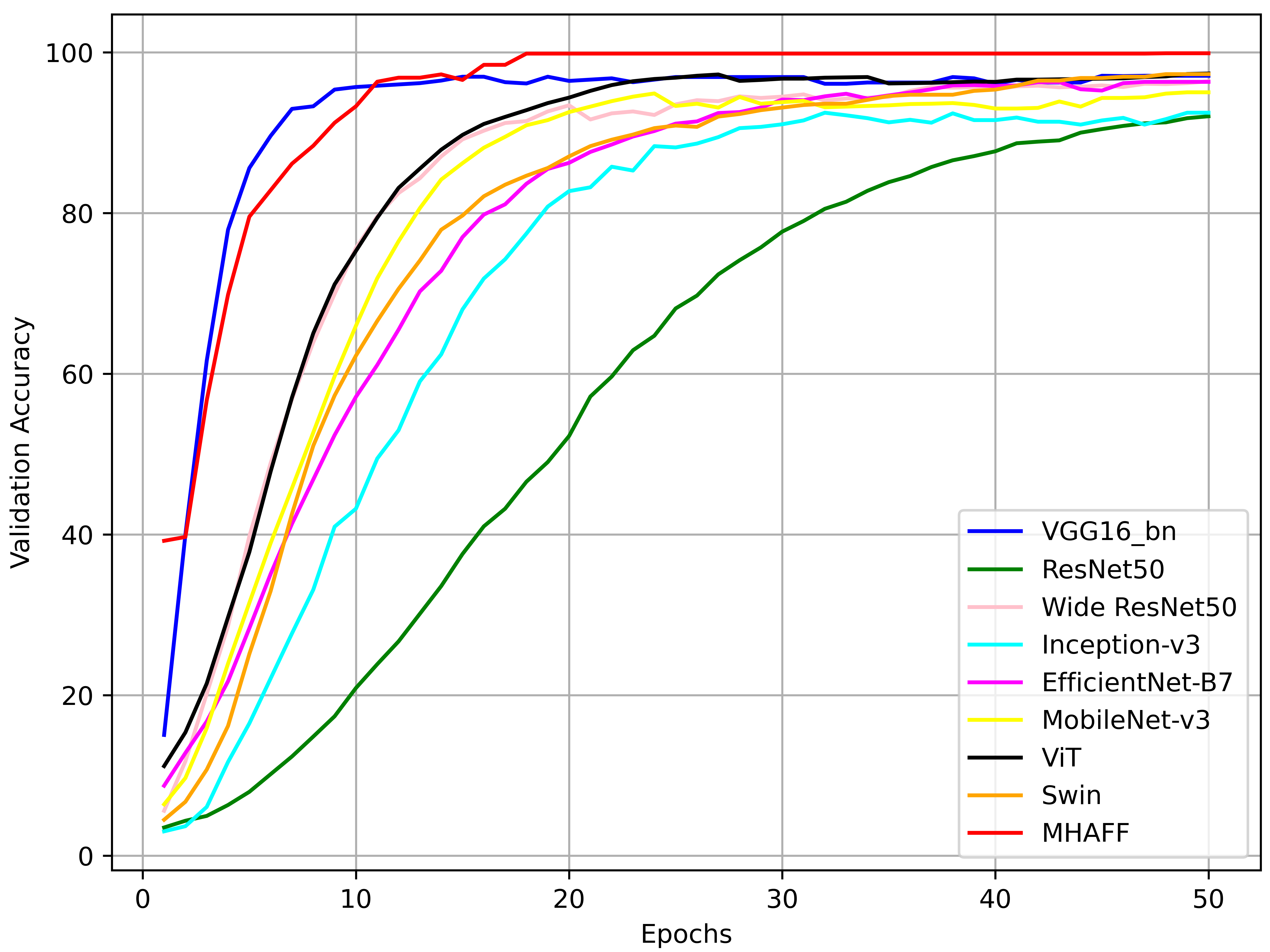}
     \caption{Validation accuracy plot for Cattle-1 with respect to epochs}
     \label{fig:acc_une}
 \end{subfigure}
 \quad
 \begin{subfigure}[b]{0.45\textwidth}
     \centering
     \includegraphics[width=\textwidth]{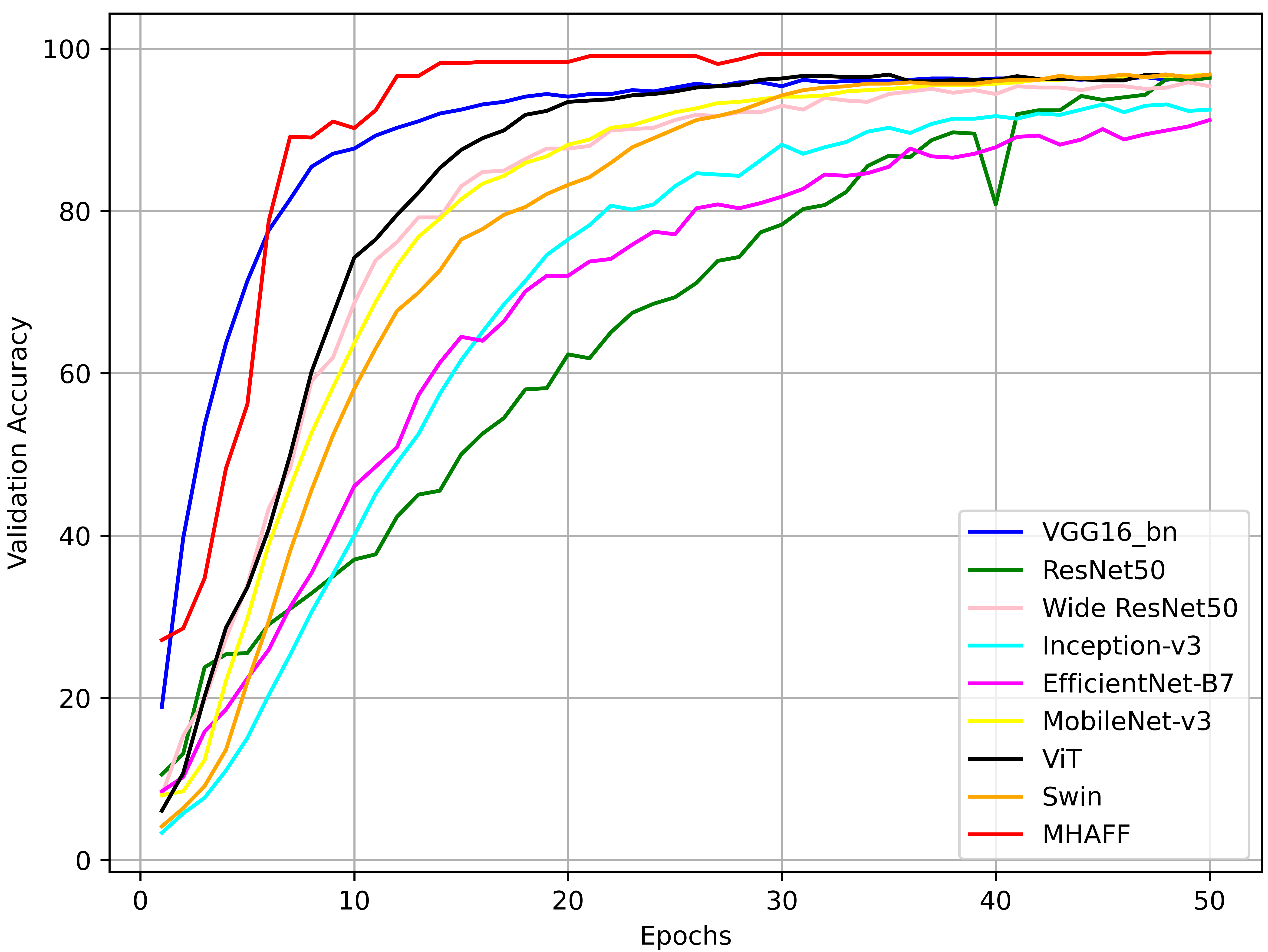}
     \caption{Validation accuracy plot for Cattle-2 with respect to epochs}
     \label{fig:acc_us}
 \end{subfigure}
 \caption{Validation accuracies of both cattle datasets}
 \label{fig:val_acc}
 \end{figure}

\begin{figure}[ht]
 \centering
 \begin{subfigure}[b]{0.45\textwidth}
     \centering
     \includegraphics[width=\textwidth]{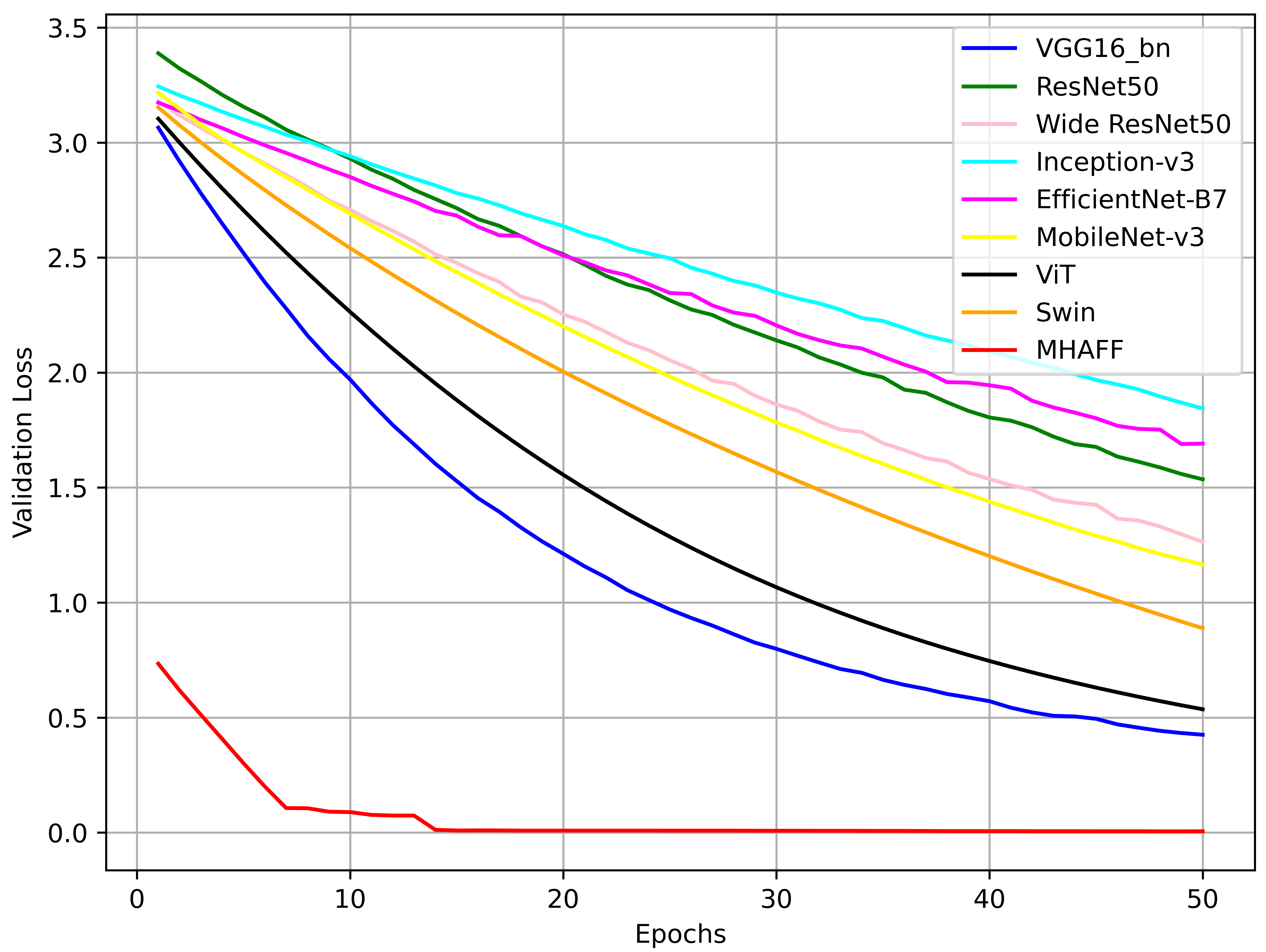}
     \caption{Validation loss plot for Cattle-1 with respect to epochs}
     \label{fig:loss_une}
 \end{subfigure}
 \quad
 \begin{subfigure}[b]{0.45\textwidth}
     \centering
     \includegraphics[width=\textwidth]{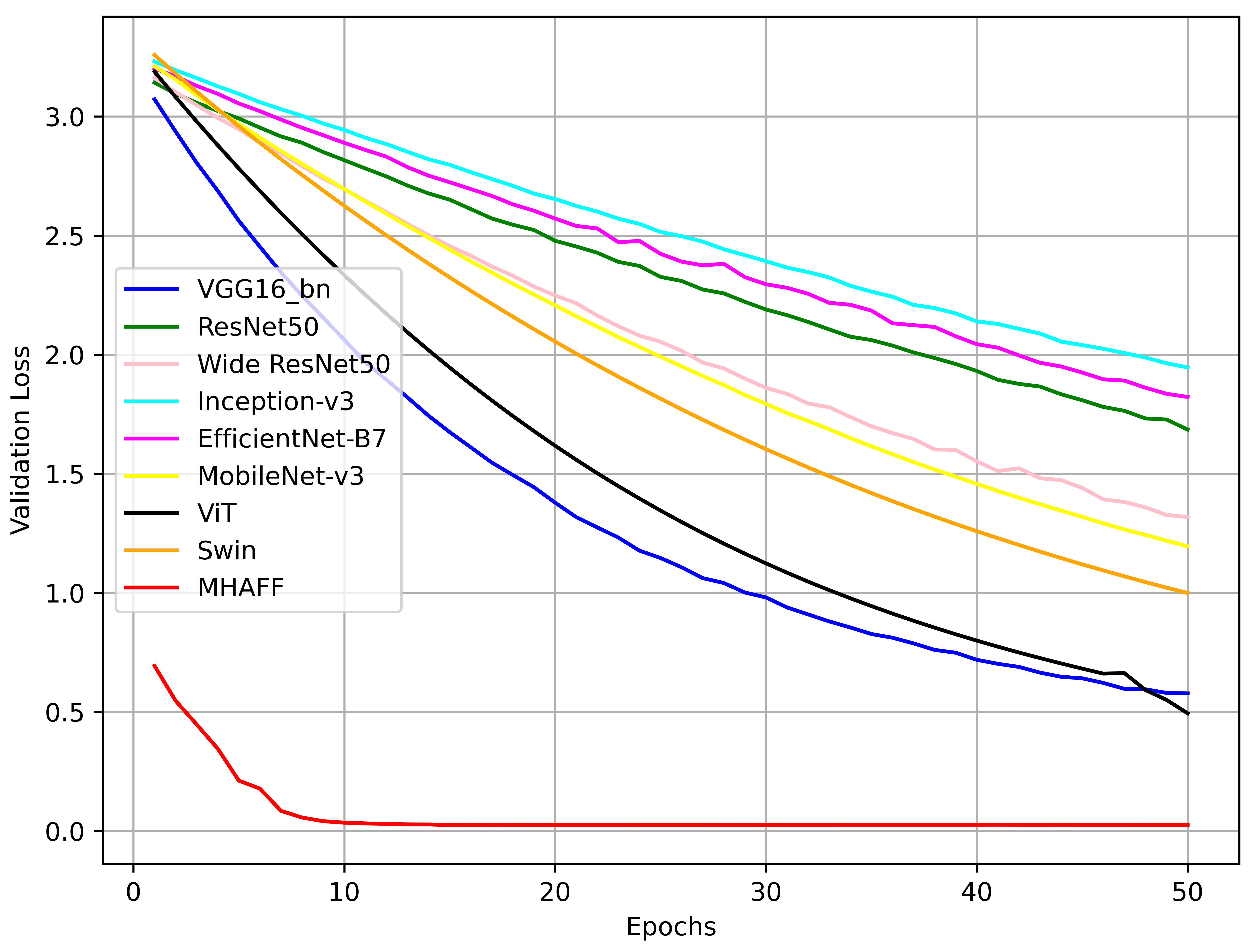}
     \caption{Validation loss plot for Cattle-2 with respect to epochs}
     \label{fig:loss_us}
 \end{subfigure}
 \caption{Validation losses of both cattle datasets}
 \label{fig:val_loss}
 \end{figure}

Table~\ref{existing} shows the highest accuracy of 99.88\% on Cattle-1 and 99.52\% on Cattle-2 data achieved by MHAFF. MobileNet-v3 achieved the second-highest performance on cattle-1 data with 97.68\%. Similarly, ViT achieved 96.80\% on cattle-2 data. MHAFF improves the accuracy by 2.2\% on cattle-1 data and 2.72\% on cattle-2.
The validation accuracy plot of the cattle identification networks is in Fig.~\ref{fig:val_acc}. The validation accuracy for the MHAFF saturates at a higher accuracy within the few epochs of the training. For instance, the accuracy of MHAFF at the 20th epoch is 99.84\% and 98.36\% for Cattle-1 and Cattle-2. On the other hand, VGG16\_bn comes in second position with 96.44\% and 94.08\%, respectively, on Cattle-1 and Cattle-2 datasets. ResNet50 comes last with 52.24\% and 62.32\% for Cattle-1 and Cattle-2 datasets. 
Furthermore, the validation loss plot of all networks is in Fig.~\ref{fig:val_loss}. The validation loss for the MHAFF network begins at a relatively low value compared to the other networks. It converges to a significantly lower loss value within a few epochs of training. For instance, the loss of MHAFF at the 20th epoch is 0.0086 and 0.0263 for Cattle-1 and Cattle-2, respectively. Following the MHAFF, VGG16\_bn achieves a loss of 1.2123 and 1.3786 on Cattle-1 and Cattle-2. The remaining methods have much higher losses at the 20th epoch. This rapid convergence suggests that MHAFF efficiently learns meaningful features from the data. The substantial difference in validation loss highlights MHAFF's ability to generalize well to unseen data, making it a robust solution for the cattle identification task. Overall, these findings underscore the effectiveness of the MHAFF approach in enhancing model performance. 

The accuracy and the loss plot diagrams show the better performance of MHAFF compared to existing cattle identification methods. MHAFF showed a fast convergence in a few training epochs with high accuracy and low loss. The superior performance of MHAFF is because of the following reasons:
\begin{itemize}
    \item Combined Strengths of CNNs and transformers: MHAFF leverages the strengths of both CNNs and transformers. CNNs excel in spatial feature extraction and pattern recognition, while transformers are adept at capturing long-range dependencies and semantic relationships across different parts of an image. By combining these strengths, MHAFF achieves a more comprehensive feature representation and understanding, improving identification accuracy. Unlike using CNN or transformer individually, MHAFF optimally utilizes spatial information and contextual understanding simultaneously. This holistic approach to representation learning ensures that all relevant features and relationships within the image are effectively captured and utilized for cattle identification. 
    \item Contextual Fusion Mechanism: MHAFF integrates a context-aware fusion mechanism through multi-head attention. This allows the capture of intricate relationships and dependencies within the image features. This capability enhances accuracy significantly in tasks where understanding spatial relationships and intricate details (like patterns of beads and ridges of the muzzle) is critical. This is proven by the Grad-CAM visualization Fig.~\ref{gradcam} as heatmaps provide the distinguishing parts of the images with sharp attention. 
\end{itemize}

\subsection{Comparison with State-of-the-art Studies}
The MHAFF was compared with the results of other cattle identification methods mentioned in the related works. Those different cattle identification methods were compared in Table~\ref{compare_other} by dataset size, number of cattle, region of interest (RoI), feature fusion, best-performing model, and accuracy. Models like ResNet50, VGG16\_bn, and EfficientNet achieve high accuracy, often above 96\%. Notably, the MHAFF model achieves the highest accuracy, 99.88\%, and 99.52\%, on two cattle datasets, outperforming all previous studies. MHAFF is validated on two cattle datasets. The source of the Cattle-1 dataset is from the research~\cite{shojaeipour2021automated}. However, we used a different approach~\cite{dulal2022automatic} to detect and extract the muzzle, resulting in the same number of cattle but a different dataset size. Notably, MHAFF achieves higher performance compared to the research~\cite{shojaeipour2021automated}. The Cattle-2 dataset is used exactly as in the research~\cite{li2022individual}, and our work surpassed the results of that research~\cite{li2022individual}. The best models used in both studies~\cite{shojaeipour2021automated, li2022individual} were also implemented in this research and compared with MHAFF in Table~\ref{existing}. However, the results vary due to different training, testing, and validation splits and differences in hyperparameter setups. Additionally, the data used in the research~\cite{shojaeipour2021automated} is not precisely identical to our Cattle-1 data.
\begin{table}[!ht]
    \centering
    \caption{Comparison between MHAFF and other research}
    \begin{tabular}{lrrlclr}
    \hline
        Study & \makecell[t c]{Dataset\\Size} & \makecell[t c]{Cattle\\Count} & RoI & \makecell[t c]{Feature\\Fusion} & \makecell[t c]{Best\\Model} & \makecell[t c]{Accuracy\\(\%)} \\ \hline\hline
         \cite{shojaeipour2021automated} & 2632 & 300 & Muzzle & No & ResNet50 & 99.11 \\
         \cite{li2022individual} & 4923 & 268 & Muzzle & No & VGG16\_bn & 98.70\\ 
         \cite{lee2023identification} & 9230 & 336 & Muzzle & No & EfficientNet & 98.10\\ 
         \cite{du2022individual} & 3153 & 34 & Trunk & Yes & Concat & 99.48 \\ 
         \cite{fu2022lightweight} & 3772 & 13 & Coat & No & ResNet50 & 98.58 \\ 
         \cite{hu2020cow} & 958 & 93 & Coat & Yes & Addition & 98.36\\ 
         \cite{li2021individual} & 3772 & 13 & Coat & No & CNN & 97.95 \\ 
         \cite{bergman2024biometric} & 7032 & 77 & Face & No & ViT & 96.30 \\ 
         \cite{zhong2023method} & 432 & 12 & Muzzle & No & Swin & 98.61\\ 
         \cite{de2020recognition} & 27849 & 51 & Coat & No & InceptionNet & 99\\ \hline
         \multirow{2}{*}{\textbf{Ours}} & 2447 & 300 & Muzzle & \multirow{2}{*}{Yes} & \multirow{2}{*}{MHA} & \textbf{99.88} \\ \cline{2-4} \cline{7-7}
          & 4923 & 268 & Muzzle &  &  & \textbf{99.52} \\ \hline
    \end{tabular}
    \label{compare_other}
\end{table}

\section{Conclusion}
In conclusion, this study introduces a novel approach to cattle identification through the Multi-Head Attention Feature Fusion (MHAFF) technique. It effectively combines the complementary strengths of Convolutional Neural Networks (CNNs) and transformers. This integration is specifically designed to address the challenges of accurate and robust identification by enhancing feature representation and focusing on crucial image regions. Through extensive experimental comparisons, MHAFF demonstrated superior performance over traditional fusion methods such as addition and concatenation, as well as existing techniques for muzzle-based cattle identification. Unlike conventional approaches, which often struggle to capture detailed and contextual information simultaneously, MHAFF’s multi-head attention mechanism allows the model to dynamically weigh and fuse features from different receptive fields. This ensures that important spatial and contextual features are prioritized, leading to a more nuanced understanding of each muzzle pattern. Grad-CAM visualizations further validated our approach, showing that MHAFF enables the model to focus on essential regions within muzzle images critical for identification. These attention maps revealed how MHAFF directs the model's focus to distinctive features. The method also demonstrated strong generalization capabilities across multiple datasets, achieving a higher accuracy rate and consistency than traditional models. By advancing cattle identification, MHAFF not only sets a new benchmark for accuracy but also represents a significant step forward in developing intelligent systems for cattle monitoring and management. The promising results of MHAFF highlight its potential for broader applications in livestock management. 
\bibliographystyle{elsarticle-num-names} 
\bibliography{cas-refs}





\end{document}